\documentclass[draftclsnofoot,onecolumn,journal]{IEEEtran}
\usepackage{cite}
\usepackage[pdftex]{graphicx}
\ifCLASSOPTIONcompsoc
  \usepackage[caption=false,font=normalsize,labelfont=sf,textfont=sf]{subfig}
\else
  \usepackage[caption=false,font=footnotesize]{subfig}
\fi
\ifCLASSOPTIONcaptionsoff
  \usepackage[nomarkers]{endfloat}
 \let\MYoriglatexcaption\caption
 \renewcommand{\caption}[2][\relax]{\MYoriglatexcaption[#2]{#2}}
\fi
\usepackage[colorlinks=True,linkcolor=blue]{hyperref}
\usepackage{amsmath}
\usepackage{verbatim}
\usepackage{lineno}
\newcommand{\m}{\textrm{m}}
\newcommand{\km}{\textrm{km}}

\begin{document}
\title{Detecting Human Interventions on the Landscape: KAZE Features, Poisson Point Processes, and a Construction Dataset}
\author{Edward Boyda, Colin McCormick, and Dan Hammer%
\thanks{E. Boyda is with the Department of Physics and Astronomy, Saint Mary's College of California, Moraga, CA 94575 USA, email: ekb2@stmarys-ca.edu}%
\thanks{C. McCormick is with Conservation X Labs, Washington, DC 20009 USA, email: cfmccormick@gmail.com}%
\thanks{D. Hammer is with the Berkeley Institute for Data Science, University of California Berkeley, Berkeley, CA 94704 USA, email: danhammer@berkeley.edu}%
}

\maketitle

\begin{abstract}

We present an algorithm capable of identifying a wide variety of human-induced change on the surface of the planet by analyzing matches between local features in time-sequenced remote sensing imagery. We evaluate feature sets, match protocols, and the statistical modeling of feature matches. With application of KAZE features, k-nearest-neighbor descriptor matching, and geometric proximity and bi-directional match consistency checks, average match rates increase more than two-fold over the previous standard. In testing our platform, we developed a small, labeled benchmark dataset expressing large-scale residential, industrial, and civic construction, along with null instances, in California between the years 2010 and 2012. On the benchmark set, our algorithm makes precise, accurate change proposals on two-thirds of scenes. Further, the detection threshold can be tuned so that all or almost all proposed detections are true positives.

\end{abstract}
\pagenumbering{gobble}
\newpage
\pagenumbering{arabic}
\section{Introduction}

The steady launch of micro- and nano- remote-sensing satellites is opening new possibilities for monitoring natural and human-driven processes on the surface of the planet. These instruments obtain imagery with resolutions of a few meters; when assembled as constellations, they are expected to offer re-imaging intervals on the order of days for any given point on Earth. Change detection at these spatial and temporal scales becomes a freshly challenging problem. 

One key feature of imagery from a constellation of lightweight satellites is that it derives from many different cameras, with individually distinct geometrical issues (e.g. lens distortions). This can lead to inconsistent registration and photometric calibration. In these circumstances, traditional pixel-based change detection techniques can no longer be relied upon. Further, pixel-based methods have always required of the analyst a certain artful, ad hoc selection of features and a decision threshold on the histogram of features. (For recent reviews and references to the extensive literature on change detection in remote sensing imagery, see \cite{ilsever12b,hussain13}.) One would like to be able, quite generally, to select any pair of time-separated images of a scene and identify locations that exhibit significant change.

Within a remote-sensing image, human-ordered structures are often set off by sharp edges, and in particular sharp corners, where pixel gradients are large in multiple directions. Local features such as those derived from David Lowe's Scale Invariant Feature Transform (SIFT), \cite{lowe99,lowe04}, are both \emph{local} and \emph{specific} to points of sharp contrast within an image. They are therefore sensitive to salient features of buildings, roads, pipelines, fields and furrows, reservoirs, clearcuts, slag heaps, and fluid discharge, to name a few examples, as well as to edges of forests, glaciers, or deserts, wherever sharp boundaries are marked on the landscape. SIFT features also offer some invariance to the position, scale, and projected geometry of features within an image. A set of relatively novel approaches to the change detection problem~\cite{gamba06,ilsever12a,dellinger14} seek to exploit these properties by matching local features across time-sequenced pairs of images.  

The idea behind local feature matching is as follows. The machine first extracts keypoints and descriptors defining local features for each of a pair of images. For every keypoint in one image, a keypoint is sought in the other image that is close to the first in both image geometry and descriptor space. Assuming stringent standards in this matching process, matched keypoints indicate a correspondence between local patches in the two images. Unmatched keypoints can result from variations in the way the images were captured, failures of the matching algorithm, or actual changes in structures on the ground. The trick, of course, is to tease apart those possibilities. This can be done by considering the keypoints in the immediate neighborhood of a given unmatched keypoint. Statistical anomalies in the matches for that set of keypoints, as compared to the matches of keypoints globally on the image, are indicative of actual, on-the-ground change.

Dellinger et al.~\cite{dellinger14} implemented this idea by modeling the appearance of SIFT keypoints as a binomial point process on the image. Given the number of matched keypoints on a neighborhood, they computed the probability that the total number of detected keypoints on the neighborhood would be greater than or equal to the number found. A low probability indicates an excess of detected as compared to matched keypoints, which can then be thresholded to give a binary change/no-change proposal for the neighborhood. In our experience, the binomial statistical model significantly outperforms a naive comparison of keypoint match rates between a neighborhood and the image as a whole.

Despite the invariances built into the SIFT features, their limitations make the change detection protocol vulnerable to variations in atmospheric conditions, in satellite and solar viewing angles, and in geometric issues with image capture. These are familiar challenges for any change detection algorithm. In this case the problem manifests as a very low global match rate between SIFT keypoints. We often observe match rates as low as a few percent, even for a pair of images from a single provider with little obvious human-interpretable change.  

When almost all keypoints remain unmatched, distinguishing those few that don't match due to on-the-ground change becomes a needle-in-the-haystack problem. The statistics don't support a reliable comparison between a given neighborhood and the rest of the image. Very often the problem lies not with matching, per se. Rather, the SIFT algorithm identifies different structures in the two images on which to extract features. In this case, relaxing the match criterion simply trades the problem of unmatched keypoints for spurious matches.  

Pursuing these observations, we offer steps to operationalize change detection in the generalized sense imagined above. The critical requirement is a high keypoint match rate in regions of the image where there is no significant on-the-ground change. We approach this challenge in three parts: (1) We introduce KAZE features~\cite{alcantarilla12} as an alternate local feature detector. Constructed through nonlinear diffusion filtering, KAZE features improve on SIFT invariances and add a measure of invariance to blurring, Gaussian noise, and nonlinear deformations of the images.   
(2) We relax the initial match criteria for keypoints, admitting as possible matches the k-nearest-neighbors in descriptor space, but then vet these proposals with a spatial proximity test and a bi-directional consistency check: a match of keypoint $(x,1)$ in image one to keypoint $(\alpha,2)$ in image two is accepted only if the top available match for keypoint  $(\alpha,2)$ is likewise keypoint $(x,1)$. Together these tactics push our average keypoint match rate above 30\%. Finally, (3) we propose a statistical test on the distribution of matched keypoints that inverts the roles of modeled and tested data from Dellinger et al. and can be expressed in terms of a marked, inhomogeneous Poisson point process on the image.  The critical advantage of this new formulation is control: By varying the probability threshold, the generation of change proposals can be made to be as permissive or selective as desired.

Although in the long term we envision applications on high-time-resolution image sequences from micro- and nano-satellite constellations, we work here with National Agriculture Imagery Program (NAIP) aerial imagery of California, predominantly from years 2010 and 2012~\cite{NAIP}. The design choice was to test the algorithm on a readily available image product, as delivered. We accessed the NAIP imagery from the public data catalog of Google Earth Engine~\cite{GEE}. Via the Earth Engine interface, the images were downsampled twice by averaging on 2x2-pixel squares, from native 1-m resolution, and served in Earth Engine standard EPSG:4326 projection. NAIP imagery comes radiometrically corrected, but calibrations are not consistent from year 2010 to year 2012. The algorithm nonetheless performs in this environment. 

As part of our contribution, we present a small benchmark dataset for change detection on this imagery, 100 pairs of images hand-selected and hand-labeled for change or lack thereof. It is a matter of interpretation as to what constitutes ``significant'' or ``interesting'' change -- the change we want to capture. As a working definition, we resolved to seek changes in structures of the sort captured by local features, concentrated in an area covering at least one percent of the image frame. The positive instances in the dataset consist of large-scale works of residential, industrial, and civic construction.  

The paper will proceed as follows. We begin in Section~\ref{sec:method} by reviewing the algorithm of Dellinger et al. based on SIFT-feature matching and the binomial statistical model. In sections~\ref{subsec:kaze}--\ref{subsec:statstests} we explain and explore the consequences of the algorithmic developments enumerated above, and in~\ref{subsec:pit} we incorporate a routine for aggregating change proposals across an image. In~\ref{sec:benchmark} we describe the genesis of our benchmark dataset. In~\ref{sec:results} we present our results: On hand-curated image pairs, we observe the challenges associated with low overall match rates and, by contrast, the consolidation of change centers and elimination of false detections which we achieve with better feature matching. On the benchmark dataset, we explore the response characteristics of the algorithm with varying change probability threshold, showing a peak overall accuracy of 68\% and a proportion of true positives among total detections that spikes to 100\%.  

\section{Methodology and Method}\label{sec:method}

We want to identify distinguishing features of a remote-sensing image and map them to corresponding elements of a later or earlier image of the same scene. Where such a mapping does not exist, we need to be able to determine if that is or is not a result of a material difference between the two scenes.  

To execute on this program, we need three components. The first is a protocol for extracting features from the two scenes. These features should capture mid-level semantics of the scene while being robust to fine-grained variations caused by differences in the image-capture process. The standard approach for image-matching applications is to use Lowe's SIFT features. However, that algorithm was developed to target invariances to the affine transformations that are common in digital photographic applications. We would instead prefer features that are robust to the particular hazards of remote sensing, including differences in illumination, solar and sensor viewing angle, lens distortion, and the sensors themselves. 

The second component is a mechanism for matching features between the two scenes. Typically this would be based on some metric of closeness on feature descriptor space, followed by a validation check such as Lowe's ratio test or a test of geometric consistency between query and target keypoints.  

The final component is a statistical model that captures the expected natural variability in matching, since perfect matching is impossible even where there is no meaningful change between two scenes. This model would be used to test for atypical distributions of matches that correspond to meaningful change. We explore each of these components in turn, taking the work of Dellinger et al.~\cite{dellinger14} as baseline.  

In their process, Dellinger et al. begin by extracting SIFT keypoints and descriptors for two time-sequenced images of a scene. They tentatively match each keypoint in one image to the keypoint in the other image with the closest descriptor. They then filter these proposals by checking against a RANSAC-estimated affine transformation between the two images. In developing a mechanism to identify change points from among the set of unmatched keypoints, the authors make two key assumptions:

\begin{enumerate}
\item Detected keypoints are distributed on the image according to a binomial spatial point process.
\item In the absence of on-the-ground change, matched keypoints are distributed with the same binomial probability of success as detected keypoints.
\end{enumerate}
We will return to examine the validity of these assumptions, but in general one would not expect a binomial or Poisson process to model the spatial distribution of keypoints. The keypoints cluster along the distinctive underlying spatial structures captured in the image, and especially outside of densely built urban areas, this leads to distributions that might better be described in terms of nodes, filaments, and voids.  

As a practical mechanism to account for inhomogeneity or interaction between keypoints, Dellinger et al. recompute the binomial probability of success locally on each neighborhood subject to the statistical test. They elect to model the matches, determining the local density of the point process by counting the number of matches $m$ on a fixed-sized neighborhood $\mathcal{N}$. Given $M$ total matches on the image, the local probability of success is $p = m/M$. This defines a binomial model $X\sim B(D,p)$ for the distribution of keypoints, of which $D$ total are detected on the image. They then test the appearance of keypoints against this model. If an anomalously large number of keypoints are detected on a neighborhood, they flag the neighborhood for change.  

Given $D$ detected keypoints on the image, $d$ detected keypoints on a neighborhood, and a hypothesized local probability of success $p$, the neighborhood is flagged as a candidate for change when the probability of finding $d$ or more keypoints is smaller than some threshold $\varepsilon$:

\begin{align}\label{eqn:pbymatchrate}
Pr_{\mathcal{N}}({X \geq d}) = \sum_{i= d}^D {D \choose i} p^i (1-p)^{D-i} < \varepsilon 
\end{align}
(In the referenced paper the probability is translated into a fraction of detected keypoints and the threshold $\varepsilon$ correspondingly rescaled by a factor of $D$.) Each unmatched keypoint defines a neighborhood $\mathcal{N}$, taken to be the ball of radius $r$ around that keypoint and truncated if necessary at image edges. The change proposal, while centered on a specific unmatched keypoint, implicates the distribution of matched and unmatched keypoints on the entire neighborhood.

The parameters $r$ and $\varepsilon$ are tuned according to the requirements of a specific application. The threshold $\varepsilon$ determines the stringency of the change criterion, while the radius $r$ defines the spatial precision of the test. For precision, one would like $r$ to be as small as possible, but of course a smaller neighborhood offers fewer keypoints on which to derive a reliable statistical model. This becomes a limiting factor especially when image-wide match rates are low. In the extreme case in which there are no matched keypoints on a neighborhood, the above test will necessarily flag the neighborhood for change - yet the lack of matches could just as well occur on a small neighborhood of an image whose keypoints are only sparsely matched due to differential effects of image capture in its image pair.  

Broadly speaking, the statistical test checks for a local deficit of matched keypoints (or corresponding excess of detected keypoints) as compared to the image-wide mean. Good global matching is essential if the local deficit is to stand out against the global match background. Because of the importance of high match rates, we devote significant attention to this issue. 

\subsection{Local Features}\label{subsec:kaze}

Inspired by Alcantarilla et al. \cite{alcantarilla12}, we examine the impact of using KAZE features instead of SIFT features for change detection. KAZE features are based on nonlinear diffusion filtering, which tends to respect edges and spatial structure more than the Gaussian scale space of SIFT. Because the imagery we are analyzing (i.e. buildings, roads, agricultural fields) is highly geometric, this seems intuitively justifiable. 

For tests of keypoint distribution and matching, we collected a corpus of image pairs from across the state of California, paired by NAIP data years 2010 and 2012. We began by specifying eight seed locations in the state, with half in heavily urban areas and half in suburban or rural areas. For each seed we obtained the 2010 and 2012 NAIP images from Google Earth Engine at 64 locations, in an 8 x 8 grid spaced at increments of 0.02 degrees latitude and 0.03 degrees longitude to the north and east, respectively, of the seed location. This produces 512 image pairs semi-randomly distributed across a variety of land cover types. Each image is 512 pixels wide by approximately 400 pixels high, varying in the north-south direction due to the EPSG:4326 projection. (Across the state of California, vertical image size ranges from 435 pixels at Tijuana to 383 pixels at the Oregon border.) Regardless, an image represents an area $2\,\km$ x $2\,\km$ on the ground, with an approximate resolution of $4\,\m$. We will refer to this set of images as our matching image corpus.

We extracted KAZE keypoints and descriptors using the OpenCV 3.2 implementation. The KAZE algorithm includes one dimensionless sensitivity parameter, called $threshold$, although this is entirely distinct from the binomial probability threshold we refer to often in the course of this paper. When using the default value for the KAZE threshold, 0.001, we found significantly fewer keypoints than with SIFT (on average only $\sim30\%$). Tuning this parameter to 0.0003, we found roughly the same number, although this varies across our matching image corpus between $\sim100$ and $\sim5000$ (median $\sim2000$). We use this parameter value for the remainder of the paper.

\begin{figure}[!thpb]
\centering
\subfloat[]{\includegraphics[width=0.5\linewidth]{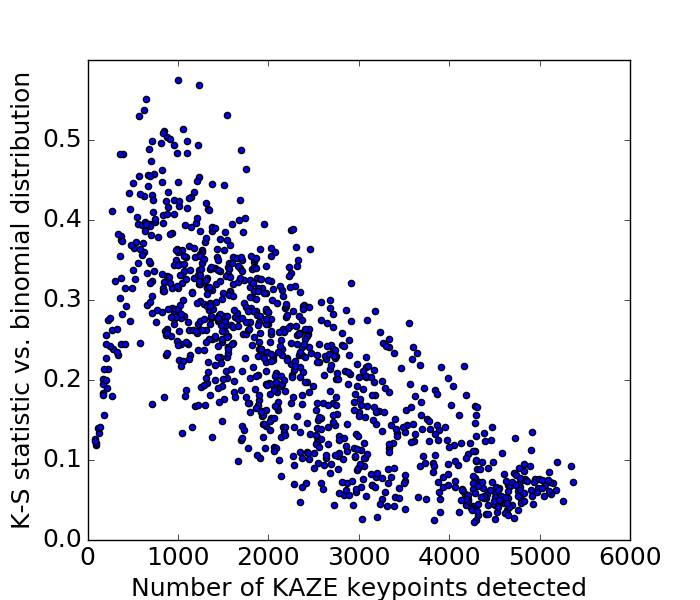}}
\hfill
\subfloat[]{\includegraphics[width=.425\linewidth]{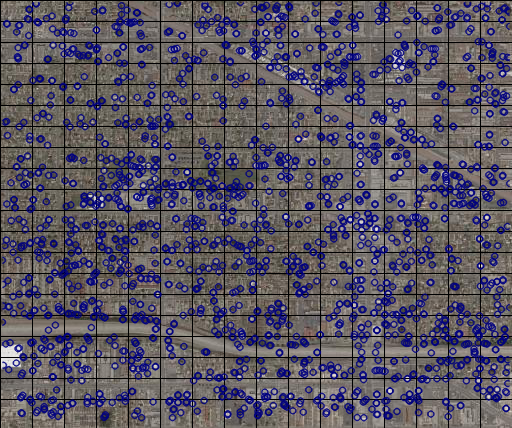}}
\subfloat[]{\includegraphics[width=.45\linewidth]{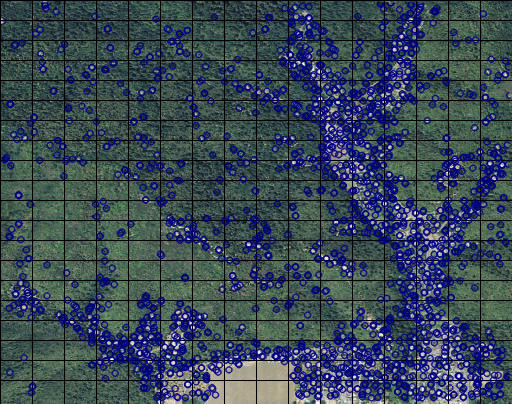}}
\hfill
\subfloat[]{\includegraphics[width=.5\linewidth]{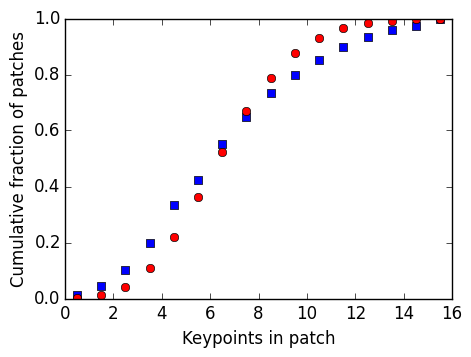}}
\subfloat[]{\includegraphics[width=.5\linewidth]{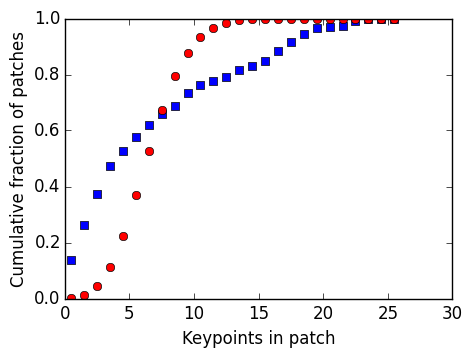}}
\caption{Tests of the spatial distribution of KAZE keypoints on the matching image corpus. (a) K-S statistic relative to binomial distribution vs. total number of keypoints. (b), (c) Images with median number of keypoints ($\sim2000$) and relatively low ($\sim0.11$ for the urban scene at left) and high ($\sim0.36$ for the mountain scene at right) K-S statistic; (d), (e) Cumulative distribution functions of keypoints per patch (blue squares) and the binomially distributed keypoints (red circles) for the images immediately above.}
\label{fig:BinomialTests}
\end{figure}

To explore the extent to which keypoints are distributed according to a binomial (or the closely related Poisson) point process, we analyzed the distribution on our matching image corpus. Using a rectangular grid of 16 x 20 neighborhoods ("patches"), we calculated the cumulative distribution function of the number of keypoints per patch and computed the Kolmogorov-Smirnov (K-S) statistic relative to a binomial distribution with the same total number of keypoints and mean number of keypoints per patch. (In some cases we discarded a small number of pixels from the edge of the image to ensure an equal number of pixels in each patch.) The resulting K-S statistic shows a large variation across our matching image corpus, with values as high as 0.5. This decreases to approximately 0.05 for images with the highest number of total keypoints. See Fig.~\ref{fig:BinomialTests}.

The lowest K-S statistics tend to occur in images with consistent land cover throughout, such as the urban scene of Fig.~\ref{fig:BinomialTests}(b), while higher values occur in scenes that include distinct land-cover boundaries, as for the  reservoir spreading its tendrils among alternately densely and sparsely tree-covered mountain slopes of Fig.~\ref{fig:BinomialTests}(c). Of particular note is the excess of squares exhibiting zero keypoints in the latter scene. The statistical model described above posits an inhomogeneous point process to account for some of this variation. Given the readily apparent filaments and voids in high K-S statistic scenes, improved modeling might focus instead on interaction among keypoints.

Finally, in passing, we observe that this analysis of keypoint distribution might be useful for classifying images into those with homogeneous versus varied land cover, although that lies beyond the scope of this paper.

\subsection{Matching}\label{subsec:matching}

As suggested by previous work, KAZE features result in significantly higher match rates than SIFT features for the same images\cite{alcantarilla12}. For matching we employed a routine from OpenCV 3.2 called BFMatcher.kNNmatch(). Given a keypoint descriptor in one image, it seeks the $k$ closest elements in descriptor space across all keypoint descriptors generated for the other image. In our initial approach, we set $k=1$. We then imposed two consistency checks:  

First, we filtered these match proposals with a geometric proximity test, discarding those that are separated in image coordinates by a distance of more than 4 pixels. We chose this value based on an estimate of the average registration offset between our images, which is approximately half this value. A hard proximity cutoff is appropriate in a context with known, bounded offsets in image registration. Establishing a fixed cutoff requires zero time and succeeds always. By contrast, generating a homography transform requires processing time and is subject to failure when based on potentially spurious nearest-descriptor-space proposals. The drawback to the proximity test is that it requires \emph{a priori} knowledge of the average image registration error. We experimented with a dynamic approach, determining a cutoff on each image from the histogram of pixel offsets for the proposals returned by BFMatcher(). Faced with a new source of imagery, we could deploy this routine in preprocessing to determine an average registration offset, but here we used our experience with NAIP imagery in fixing a 4-pixel cut-off. 

Second, we required that matches be symmetric, such that if the proposed match for keypoint $(x,1)$ in image one is keypoint $(\alpha,2)$ in image two, then the proposed match for keypoint $(\alpha,2)$ is likewise keypoint $(x,1)$. This cross-check expresses a logically necessary condition for two keypoints to match, but it is not met by all proposals which pass the proximity test; we found that approximately $15\%$ of proposed matches fail this test for our matching image corpus.

We define the match rate for a pair of images as twice the number of successful matches divided by the sum of the total number of keypoints detected in each image. Using SIFT features on our matching image corpus, the average match rate was $11.8\%$, with a small positive correlation with total keypoint number. When we relaxed the proximity test to 8 pixels, the average match rate increased to $12.2\%$; when we tightened it to 2 pixels, the average match rate decreased to $10.3\%$. With KAZE features and the 4-pixel proximity cut-off, the average match rate was $26.8\%$, 2.3 times higher than with SIFT.

\begin{figure}[!thpb]
\centering
\includegraphics[width=.5\linewidth]{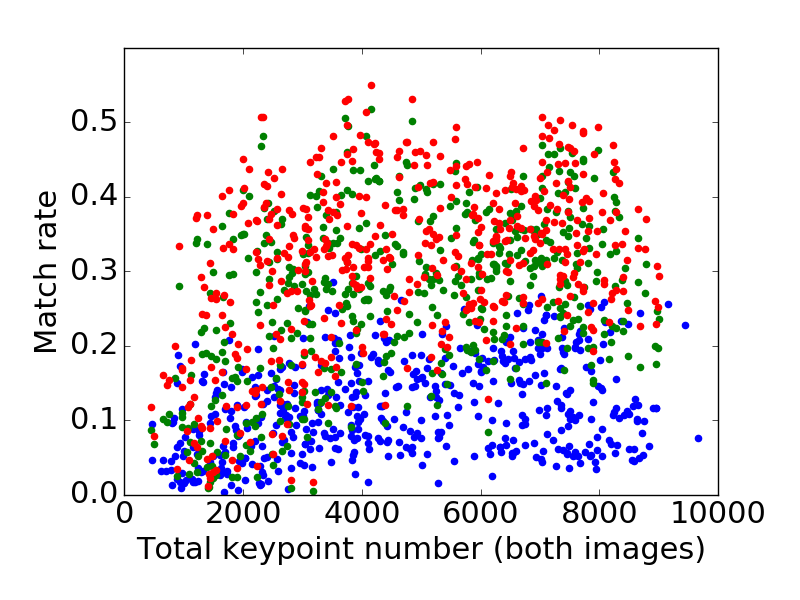}
\caption{Match rates across the matching image corpus for different matching procedures, all of which employ the OpenCV function BFMatcher.knnMatch(), a fixed 4-pixel proximity test, and the forward-backward cross-check described in the text. Blue: SIFT features. Green: KAZE features. Red: KAZE features with kNN matching ($k=5$).}
\label{fig:MatchRates}
\end{figure}

To improve our match rates, we continued to use KAZE features and expanded the parameters of the initial search performed by BFMatcher.kNNmatch() to return the top $k$ nearest neighbors in descriptor space, typically with $k=5$. In $83\%$ of cases the nearest neighbor passed the (4-pixel) proximity test; in the remaining cases we searched for the next-closest match that passed. We repeated this search procedure, reversing the role of the two images, and performed the cross-check described above. This approach increased the average match rate to $31.1\%$. Results of these match tests are summarized in Fig.~\ref{fig:MatchRates} and in Table \ref{table:MatchingApproaches}.

\begin{table}[!t]
\centering
\caption{Summary of matching approaches and average match rates.}
\begin{tabular}{|l|c|}
\hline
SIFT, 4-pixel proximity text, cross-check & $11.8\%$ \\ \hline
KAZE, 4-pixel proximity test, cross-check & $26.8\%$ \\ \hline
KAZE, kNN matching ($k=5$), 4-pixel proximity test, cross-check & $31.1\%$ \\
\hline
\end{tabular}
\label{table:MatchingApproaches}
\end{table}

\subsection{Statistical tests}\label{subsec:statstests}

The idea behind change detection based on local feature matching is that on-the-ground change ought to be reflected in an anomalously low match rate among features in a corresponding region of an image. An immediate way to operationalize this approach would be to threshold the match rate on a neighborhood as compared to the average match rate on the image, which amounts to postulating and thresholding a triangular probability distribution for matched keypoints. However, this test does not give consistent results: A threshold tuned to yield reasonable detections for one scene results in significant false positives or negatives in others. 

Alternately, one could consider the matched keypoints to be uniformly distributed among total keypoints on the neighborhood and threshold the binomial probability that the given number of matches appear. We assign the same meaning to variables as above. If there are $M$ matches from $D$ total detected keypoints on the image, the probability that a given keypoint is matched is $p' = M/D$.  This gives rise to the binomial distribution $X' \sim B(d,p')$ of matches among the $d$ detected keypoints on the neighborhood $\mathcal{N}$. If $m$ matches out of $d$ keypoints are observed on the neighborhood, the change criterion is

\begin{align}\label{eqn:matchkpratio}
Pr_{\mathcal{N}}({X' \leq m}) = \sum_{i= 0}^m {d \choose i} (p')^i (1-p')^{d-i} < \varepsilon'
\end{align}
This approach yields results qualitatively similar to our preferred statistical test, equation~(\ref{eqn:pbykprate}) below. We will return to discuss the close mathematical relationship between them.

\begin{figure}[!thpb]
 \centering
\includegraphics[width=\textwidth]{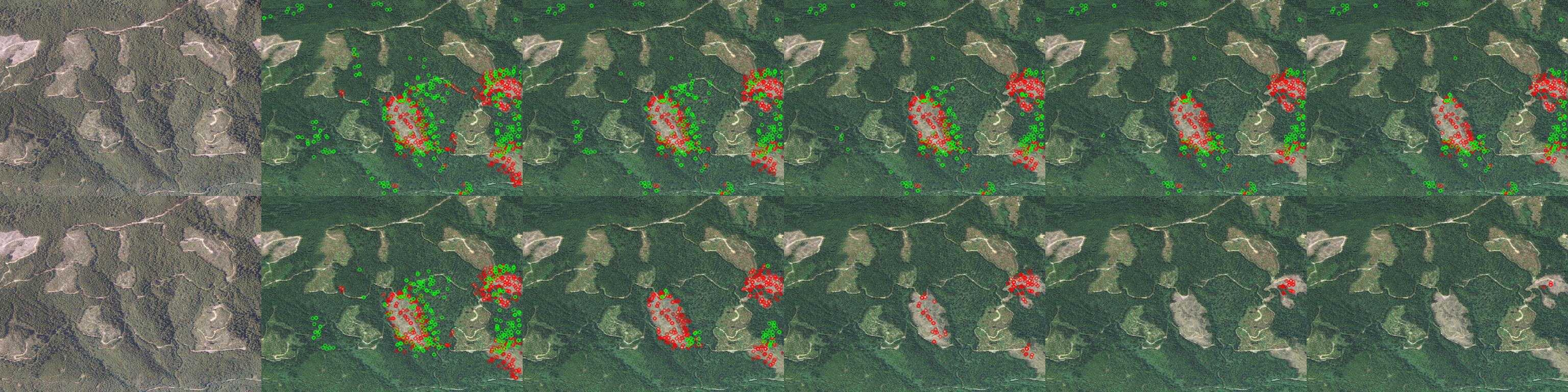}
\caption{Tuning down the probability threshold, on clearcuts near Crescent City, CA, 2010-2012. Top, using statistical test~(\ref{eqn:pbymatchrate}) and probability thresholds $\log\varepsilon \in  [-8,-40]$, proposed change points persist despite the progressively smaller thresholds.  Bottom, with test~(\ref{eqn:pbykprate}) and $\log\tilde\varepsilon \in [-2,-10]$, the change points settle within the fresh clearcuts and eventually disappear. Far left is the reference image from 2010.}
\label{fig:turndownepsilon}
\end{figure}

Dellinger et al. postulate a binomial point process on the image, electing to model the matched keypoints and subsequently test the distribution of all keypoints against this model. Intuitively, we prefer the inverse, to model the keypoints and test the distribution of matches. We thus derive a (third, distinct) binomial distribution $\tilde X\sim B(M,\tilde p)$, with probability of success dictated by the local distribution of detected keypoints, $\tilde p = d/D$. A change proposal is generated when the probability of finding $m$ or fewer matches on the neighborhood, out of $M$ total matches on the image, is less than a threshold $\tilde \varepsilon$:

\begin{align}\label{eqn:pbykprate}
Pr_{\mathcal{N}}({\tilde X \leq m}) = \sum_{i= 0}^m {M \choose i} (\tilde p)^i (1-\tilde p)^{M-i} < \tilde \varepsilon
\end{align}
In testing for the appearance of anomalously few matches, as opposed to anomalously many keypoints, the survival function of~(\ref{eqn:pbymatchrate}) is replaced here by the cumulative distribution function. The critical practical advantage offered by~(\ref{eqn:pbykprate}) as compared to~(\ref{eqn:pbymatchrate}) is its sensitivity to the threshold $\tilde \varepsilon$.  For instance, to eliminate false positives at the expense of missed detections, we would want to be able to tune down the number of proposed change points by tightening the threshold. In the case that there are, for example, zero, one, two, or three matched keypoints on $\mathcal{N}$, as often happens, the model for~(\ref{eqn:pbymatchrate}) is derived from those very few data points.  The probabilities generated are zero or near zero, at the limits of machine precision. Such neighborhoods are always flagged for change, for any viable threshold $\varepsilon$. On the other hand, the test~(\ref{eqn:pbykprate}) responds readily to a reduction in threshold. The probability distribution is derived from the more numerous keypoints, and the cumulative distribution function outputs probabilities that are non-zero and finite even when there are zero matches on a neighborhood.  See Fig.~\ref{fig:turndownepsilon}.

Although a comprehensive analysis of the possibilities of statistical modeling lies beyond the scope of this paper, we offer the following observations.  The models~(\ref{eqn:matchkpratio}), (\ref{eqn:pbykprate}) can be viewed as different conditionings of a marked, inhomogeneous Poisson point process on the image, in which the marks $\mu \in \{1,0\}$ indicate whether a keypoint is or is not matched, assumed uniformly distributed among the keypoints, and the inhomogeneity accounts for the local variations in intensity of the Poisson process.  Conditioning on the match criterion ($\mu =1$) and a total number of $M$ matches on the image projects to the distribution of a binomial point process~(\ref{eqn:pbykprate}). By contrast, conditioning on the observed number of keypoints $d$ in a given neighborhood $\mathcal{N}$ projects to a binomial distribution of marks (i.e. matched keypoints) among keypoints on the neighborhood. This is~(\ref{eqn:matchkpratio}). The latter operation involves conditioning the distribution of the marked, inhomogeneous point process for each neighborhood subject to the statistical test. (For an introduction to spatial point processes, see, e.g.~\cite{baddeley04}.)

Generically, as noted above and as seen in Fig.~\ref{fig:BinomialTests}, the keypoints are not Poisson distributed on the image. It would be interesting, as a direction for future exploration, to see what improvements in change detection might follow from modeling statistical interactions among the keypoints.

\subsection{Putting it together}\label{subsec:pit}

Proposed change points often cluster in regions exhibiting the most significant change. At the same time, we frequently observe a few scattered change points in regions which, subjectively, we would not associate with meaningful change. We wish to leverage these observed patterns to make summary binary change / no change assertions for our image pairs. This being a task in pattern recognition, it would seem ripe for a supervised approach. Here we opted simply to count change points in local neighborhoods and assert change where this count exceeds some fraction of the average number of keypoints on the two images.  

There is an asymmetry in the ordering of the images in a pair. In cases where new structures appear on a previously blank landscape, for instance in building construction on an empty lot, few keypoints will be extracted in the blank regions of the initial image, whereas many may be extracted on the new structure in the second image. The keypoints in the second image will be unmatched and flagged for change. In the first image there are few or no keypoints to consider, and the blank region won't be flagged for change. The opposite occurs when structures disappear. Clearly, to capture change in both scenarios we need to check in both directions, \emph{forward}, considering matches for keypoints from the earlier image onto keypoints from the later image, and \emph{backward}, considering matches for keypoints from the later image onto keypoints from the earlier one.

To make final change / no change proposals, we sum both forward and backward change points on the image plane by convolution with a uniform square kernel and then threshold the output, thus aggregating change points into smooth change regions, or windows. With limited testing we were able to determine workable settings for the neighborhood size and threshold. The key steps of the algorithm, along with our preferred parameters for the Earth Engine-derived NAIP imagery, are given in Table~\ref{tab:params}.

\begin{table}[!thpb]
\centering
\caption{Parameters of the algorithm tuned for NAIP imagery at $\sim \,$4-m resolution.}
\hspace*{-8mm}
\begin{tabular}{|l|l|}
\hline
Function & Parameters \\
\hline  
1. KAZE keypoint extraction & Sensitivity threshold: .0003\\
2. Matching on $k$ nearest descriptors subject to proximity test + cross-check & $k = 5$; Radius: 4 pixels  \\
3. Binomial statistical test~(\ref{eqn:pbykprate}) & Threshold $\tilde\varepsilon \in [10^{-4},10^{-8}]$; Radius($\mathcal{N}$): 30 pixels \\ 
4. Change point aggregation & Window: 120 x 120 pixels; Threshold fraction: 0.1\\
\hline
\end{tabular}
\label{tab:params}
\end{table}

\section{Benchmark Dataset}\label{sec:benchmark}

We set out to detect instances of concerted human activity on the landscape. The algorithm as presented is sensitive to construction and demolition, clearing and regrading of the land, shifts in agricultural use, and natural processes that impact sharp divisions in land-cover classes.  At the same time, we want to avoid detecting seasonal vegetative change and small-scale, transitory elements such as street traffic. The division between change of interest to human operators and all other differences between images is necessarily subjective.  We still have not settled on a viable distinction in agricultural contexts.  Construction, however, is relatively straightforward to observe and define.  For this reason, and for its inherent interest, we made construction the basis of our benchmark dataset.

We established a minimal scale to filter small, scattered elements from consideration: We require that our target change regions cover at least one percent of the image frame.  At our current working resolution, that means a change region, if square, extends minimally two hundred meters on a side. Clearly we are excluding some objects of potential interest, such as single-home construction.  In actuality, the smallest region we included in the benchmark dataset covers 1.6\% of the area of the image, and even this may have been too small:  Our algorithm struggles to detect these smallest instances, given the parameter settings required to limit false-positive detections.  

The bulk of our examples we found by hand-search of imagery in Google Earth, supplemented by news reports of construction from 2011. From Earth Engine we downloaded the scene for years 2010 and 2012 containing the construction of interest, often off-centered so as to avoid any secondary confounding instances within the frame.  We then scrolled until we found a disjoint, nearby (no more than a tenth of a degree distant) scene in which any observable change was smaller in extent than our minimal allowed scale.  In this way we built a corpus of fifty positive and fifty negative sample image pairs, from predominantly urban and surburban contexts, but including also some examples with surrounding farmed fields, desert, and green space. 

For the fifty pairs exhibiting change, the change region covered on average $7\pm 5\%$ of the image frame.  The maximum coverage was 31\% and the minimum, aforementioned, 1.6\%.  The construction examples fall into the following rough categories: twelve housing subdivisions, seven box stores or malls, five parks or athletic facilities, four hospitals, seven schools, two prisons, two cases of highways or roads, five varied industrial / warehouse / office park facilities, and six cases of demolition or site clearing.

We drew polygonal bounding boxes around areas of construction with an image editing tool. On testing, we considered any intersection of a hand-drawn bounding box with an output aggregated change region to constitute a true positive detection for the scene. The benchmark dataset is available on GitHub~\cite{githubbench}.

\section{Results}\label{sec:results}

We begin with some qualitative observations.  The matching design factors most impacting the accuracy of change detection are the feature set (SIFT vs. KAZE) and the $k$-nearest-neighbor matching protocol.  These variants are shown in Figs.~\ref{fig:ConstructionTiles} and~\ref{fig:CCBBTiles} below.  The original images from 2010 and 2012 appear first, and the sequence below shows keypoints flagged for change under the following four scenarios:  SIFT features; SIFT features with kNN matching ($k=5$); KAZE features; KAZE features with kNN matching ($k=5$). The scenes were hand-selected and are separate from the benchmark dataset.  For purposes of this demonstration, the probability thresholds $\tilde\varepsilon$ were determined to yield roughly comparable numbers of change points on the first of the Fig.~\ref{fig:ConstructionTiles} construction scenes. 

\begin{figure}[!thpb]
\centering
\subfloat{\includegraphics[width=.8\textwidth]{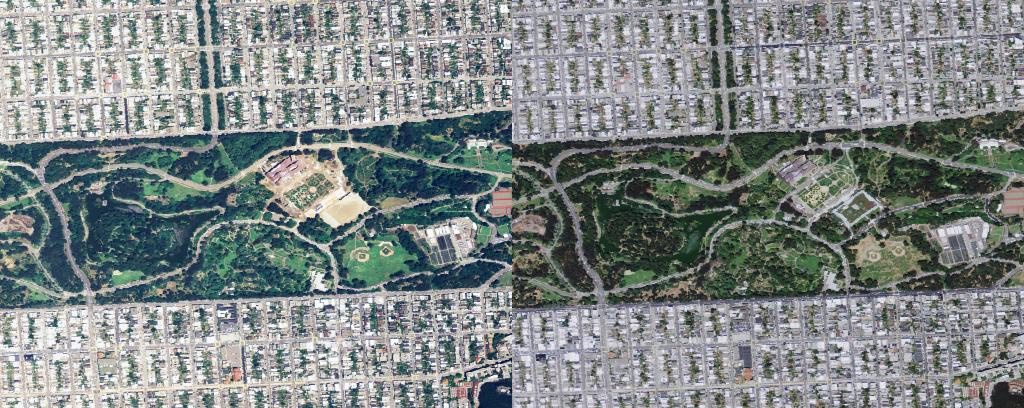}}
\hfill
\setcounter{subfigure}{0}
\subfloat[]{\includegraphics[width=\textwidth]{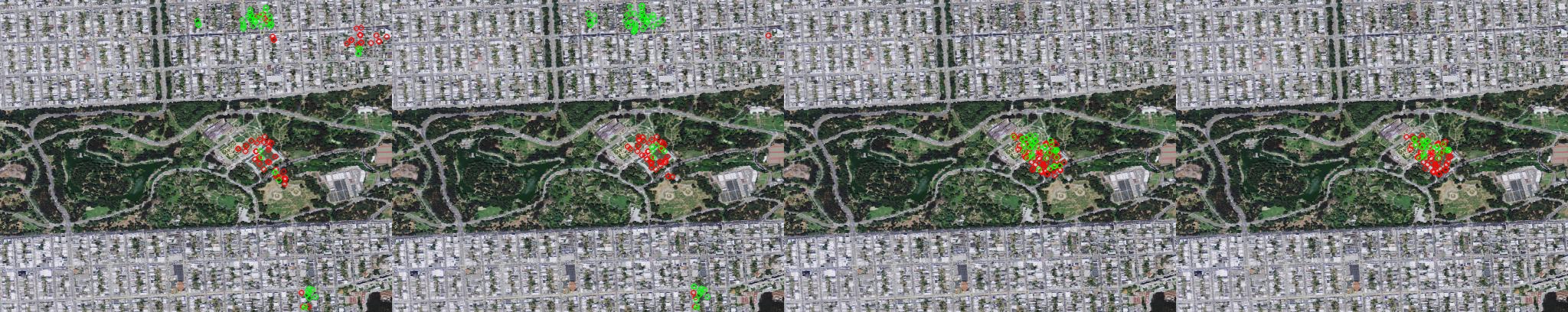}}
\hfill
\subfloat{\includegraphics[width=.8\textwidth]{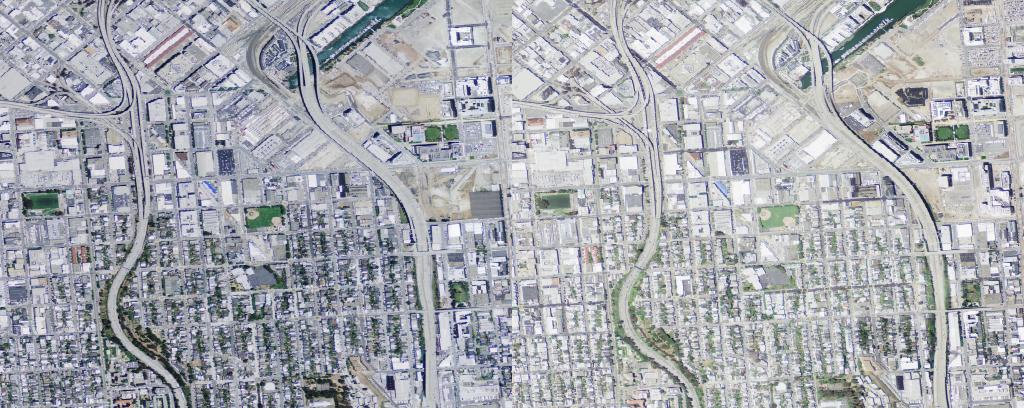}}
\hfill
\setcounter{subfigure}{1}
\subfloat[]{\includegraphics[width=\textwidth]{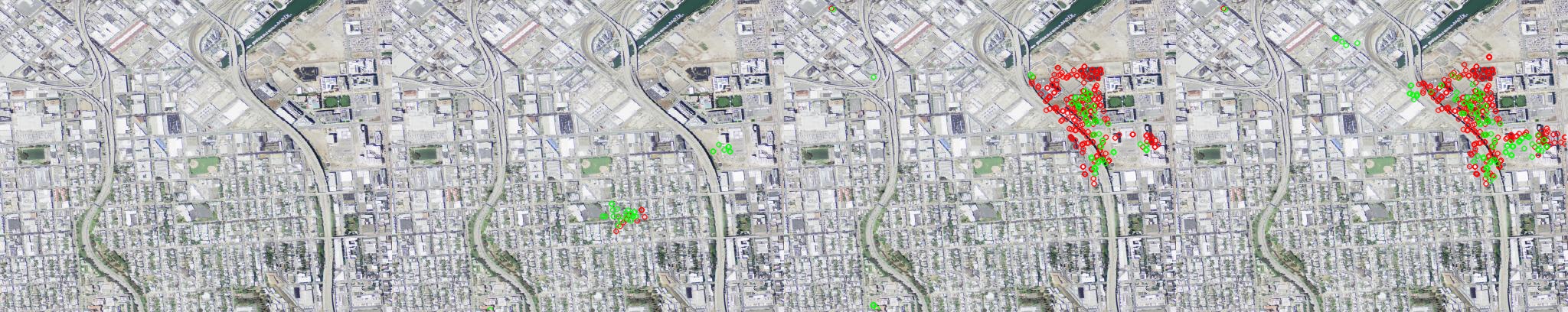}}
\caption{Input image pairs and output sequences showing successive adjustments to matching protocols. (a) Construction of the California Academy of Sciences, San Francisco, 2005-2009. (b) Construction UCSF Mission Bay campus, San Francisco, 2009-2012. From left to right the output sequences were generated using: SIFT features; SIFT with kNN matching ($k=5$); KAZE features; KAZE with kNN matching ($k=5$).  Thresholds for the statistical test~(\ref{eqn:pbykprate}) were set (SIFT: $\tilde\varepsilon = 10^{-3}$; KAZE: $\tilde\varepsilon = 10^{-6}$) for illustrative purposes and so as to yield roughly comparable numbers of detected change points across the first sequence (in order: 111, 106, 112, 87 change points). Backward and forward change point proposals appear, respectively, in green and red.} 
\label{fig:ConstructionTiles}
\end{figure}

\begin{figure}[!thpb]
\centering
\subfloat{\includegraphics[width=.8\textwidth]{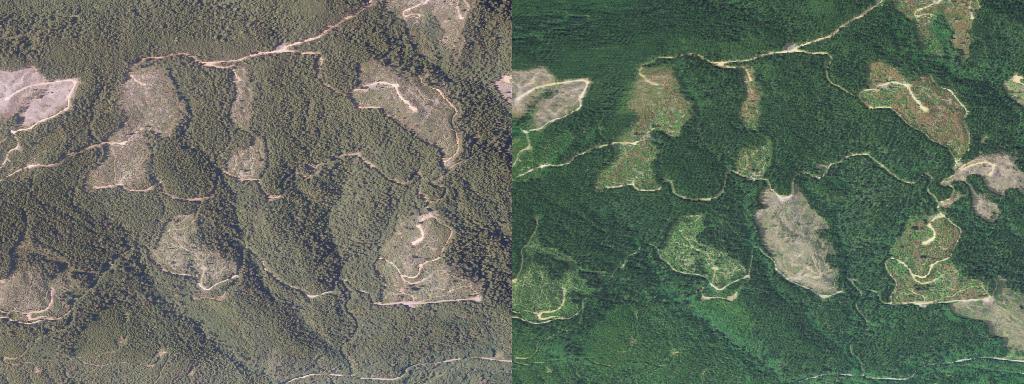}}
\hfill
\setcounter{subfigure}{0}
\subfloat[]{\includegraphics[width=\textwidth]{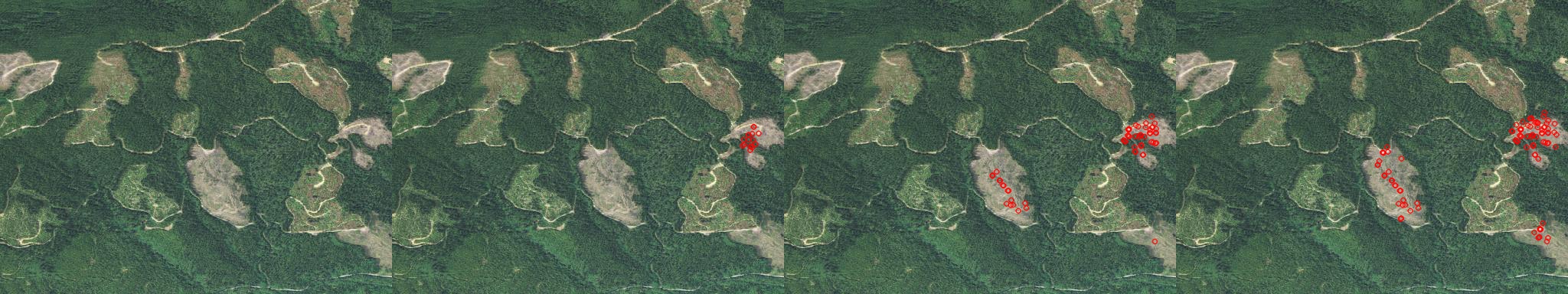}}
\hfill
\subfloat{\includegraphics[width=.8\textwidth]{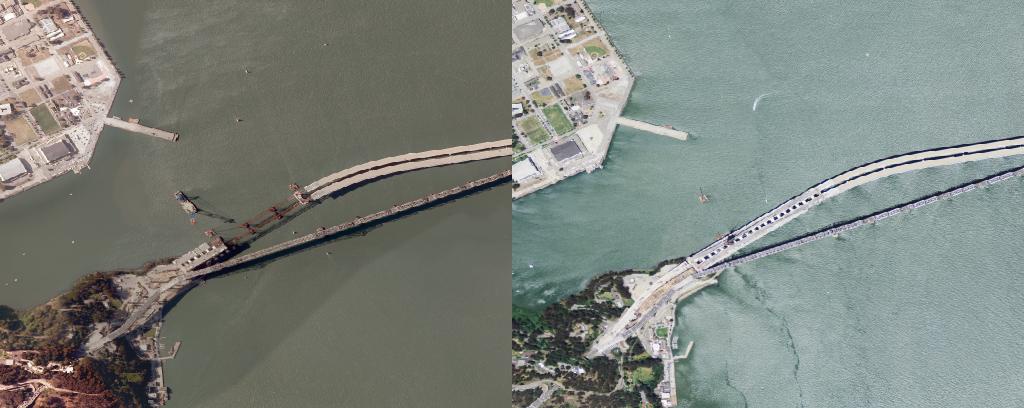}}
\hfill
\setcounter{subfigure}{1}
\subfloat[]{\includegraphics[width=\textwidth]{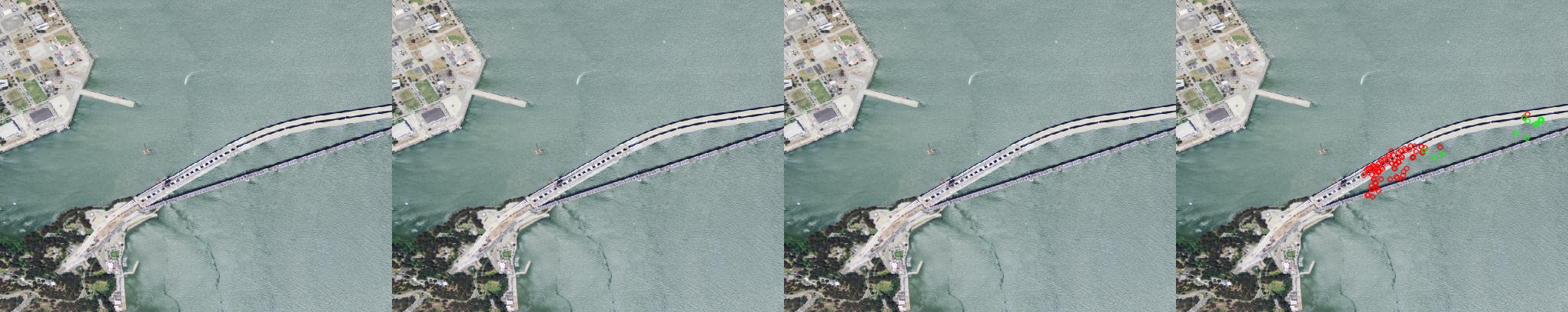}}
\caption{Applying the same test sequence and thresholds as in Fig.~\ref{fig:ConstructionTiles}, the SIFT-based routines (left and second-from-left) fail to detect readily apparent change in these difficult-to-match scenes. SIFT-based match rates are only 4-7\%. (a) Clearcuts northeast of Crescent City, CA, 2010-2012.  (b) Construction of the San Francisco Bay Bridge, 2010-2012. In the Bay Bridge scene, KAZE features with kNN matching ($k=5$) increases the number of matched keypoints to 259 from 171 with KAZE alone.}
\label{fig:CCBBTiles}
\end{figure}

For all variants in the first sequence, the construction of museum and plaza at center of the frame is correctly identified.  However, the SIFT-based routines propose more scattered change points in areas with no obvious change. In the second sequence, the new buildings on the UCSF Mission Bay campus, upper right, are peppered with KAZE change points but are almost entirely missed when using SIFT.  
In Fig.~\ref{fig:CCBBTiles}, maintaining the same probability thresholds as for Fig.~\ref{fig:ConstructionTiles}, KAZE with kNN matching zeros in on the three clearcuts and the Bay Bridge construction, while SIFT-based routines largely fail to capture the notable change.  The mottled, shifting faces of forest and bay make for challenging matching tasks, and the overall SIFT match rates on these image pairs are in the range of 4-7\%, as compared to 10-23\% for KAZE.  By relaxing the probability threshold, the SIFT cases can capture the change, but this comes at the cost of significant false-positive detections.

\begin{figure}[!thpb]
 \centering
\subfloat[]{\includegraphics[width=\textwidth]{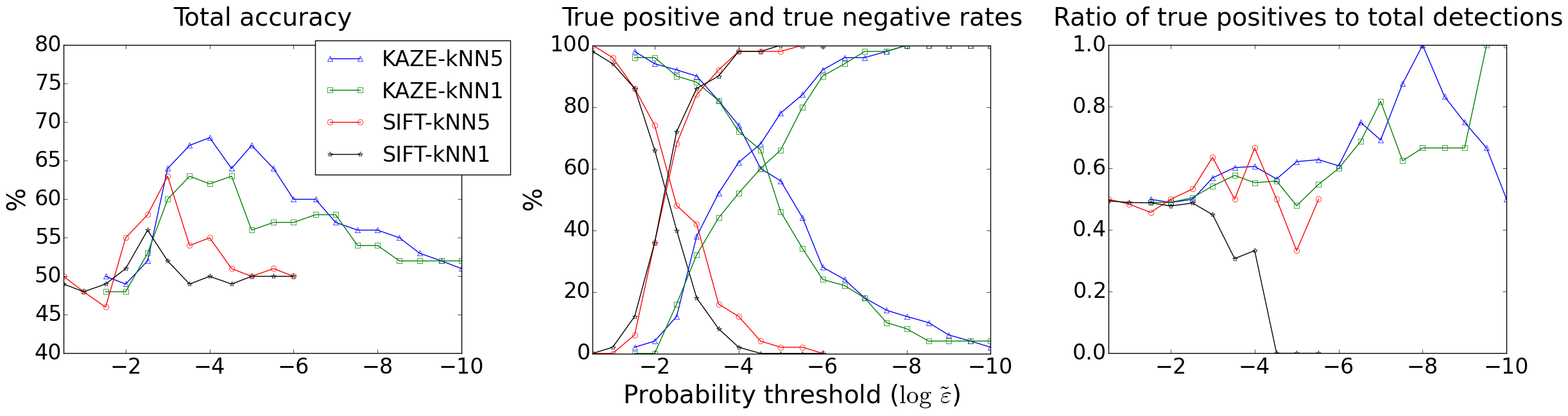}}
\hfill
\subfloat[]{\includegraphics[width=\textwidth]{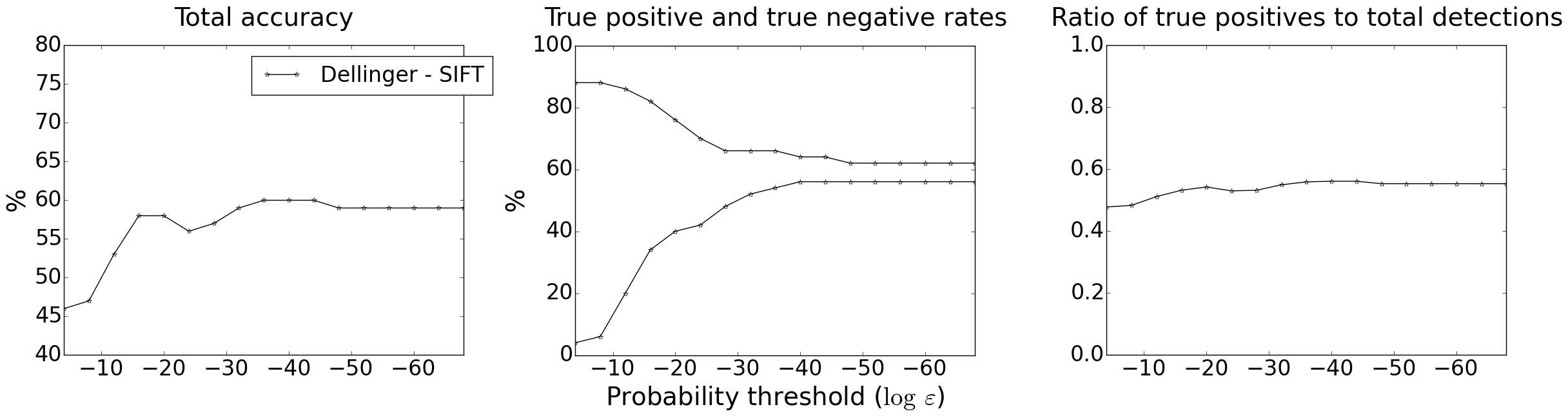}}
\caption{Results on the test dataset for different versions of the change algorithm. (a) Our full program as summarized in Table~\ref{tab:params}, plus variants for feature set (SIFT, KAZE) and kNN matching.  Black and red curves terminate where total detections go to zero.  (b) The original program of Dellinger et al.~\cite{dellinger14}, using SIFT features, homography for match verification, and the statistical test~(\ref{eqn:pbymatchrate}), with the addition of our aggregation routines.  The leftmost plots indicate total accuracy, which peaks at 68\% for the program of Table~\ref{tab:params}.  Center plots indicate dependence of true positive rates (decreasing) and true negative rates (increasing) on decreasing probability threshold. At top, true positive and true negative rates saturate for appropriate values of the threshold, while at bottom, the threshold ceases to affect output beyond $\log\varepsilon=-48$. The rightmost plots demonstrate a potential filtering application: Our program can be tuned to yield a high proportion of true positives among total detections.}
\label{fig:accuracies}
\end{figure}

We reserved the benchmark dataset for the following quantitative tests, and did not experiment with it prior to this point. The graphs of Fig.~\ref{fig:accuracies} show data for several versions of the change algorithm as applied to this dataset. In the plots top-left, a progression can again be seen as improved matching techniques are introduced, from a best accuracy of 56\% with SIFT alone, up to an overall optimal accuracy 68\% with KAZE features and kNN ($k=5$) matching. Blanket change or no-change assertions across images would yield 50\% accuracies, but as our program yields compact, targeted proposals, it could have fared much worse: For the optimal settings, at threshold $\tilde\varepsilon = 10^{-4}$, the average fractional area of a proposed change window is $12\pm 7\%$; at $\tilde\varepsilon = 10^{-6}$ this drops to $8\pm 5\%$. Meanwhile, recall, the average fractional area of a labeled change region is $7\pm 5\%$. An erroneous proposed detection and a labeled change region need not intersect. 

With the statistical test~(\ref{eqn:pbykprate}), reducing the probability threshold reduces change proposals such that true positive and true negative rates both exhibit sigmoidal dependence on the logarithm of the threshold. As shown top-center, near 100\% true positive rates become 100\% true negative rates as the threshold tightens by six orders of magnitude.  This responsiveness enables the user to tune the algorithm for different applications. To monitor a given location, one might set the threshold so as to capture any possible event, even at the cost of false positive detections.  On the other hand, to filter large bodies of satellite-derived imagery to identify scenes for further, possibly human, interpretation, limiting false positives would be a priority. Here the qualities of our program particularly stand out.  At restrictive thresholds it returns a high proportion of true positives out of total detections. For $\tilde\varepsilon\in \{10^{-6},3\cdot 10^{-7},10^{-7},3\cdot 10^{-8}, 10^{-8}\}$, the program hits 14 of 23, 12 of 16, 9 of 13, 7 of 8, and 6 of 6, true positives out of total change proposals, respectively. Full data for the program variants' proportions of true positive detections are plotted in the rightmost panel of the triptych.  

Data output by the program of Dellinger et al., using SIFT features, homography for match verification, and the statistical test~(\ref{eqn:pbymatchrate}), with the addition of our aggregation routines, are plotted in the bottom triptych.  The overall best accuracy attained is 60\%, somewhat lower than the 68\% we achieve, although perhaps more notable is the way the accuracy remains nearly constant as the threshold tends to an infinitessimal value. As discussed in section~\ref{subsec:statstests}, and readily visible in the graph bottom-center of Fig.~\ref{fig:accuracies}, under test~(\ref{eqn:pbymatchrate}) many keypoints are flagged for change independent of the threshold $\varepsilon$.  The baseline accuracy of  59-60\% seen in the lower triptych depends on the assertion of change on large swathes of many of the scenes, capturing by default many instances labeled for change. At $\varepsilon = 10^{-36}$, where 60\% accuracy is attained, the average area of a proposed change window is $33\pm 32\%$. No matter how small the threshold, false positive rates never decline below 44\%.  As a related fact, seen bottom-right, the fraction of true positives of total detections never exceeds 55\%.

\begin{figure}
\centering
\subfloat[]{\includegraphics[width=\textwidth]{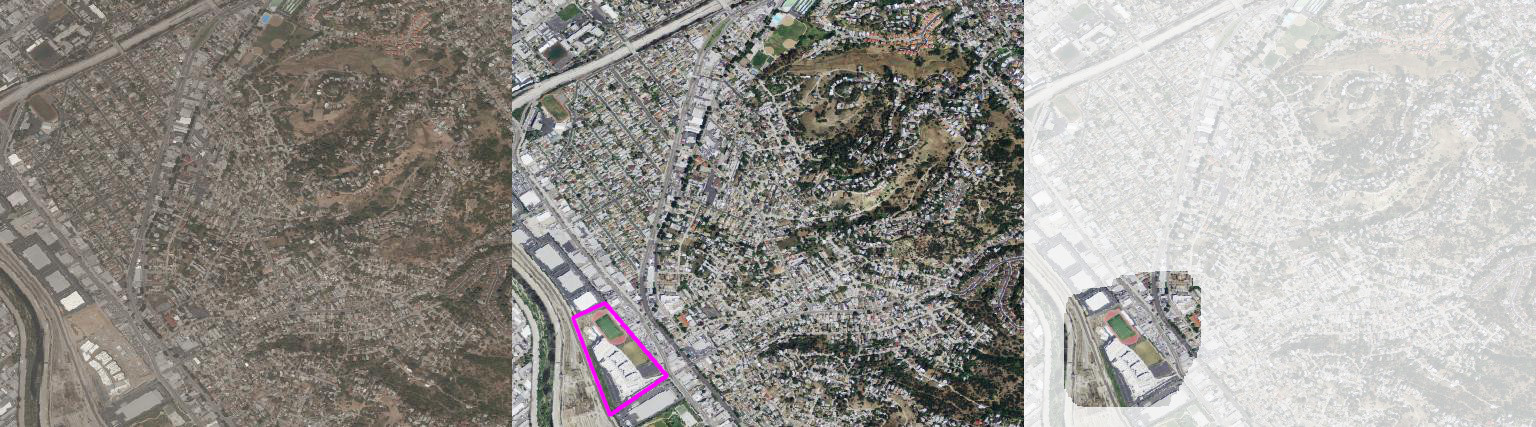}}
\hfill
\subfloat[]{\includegraphics[width=\textwidth]{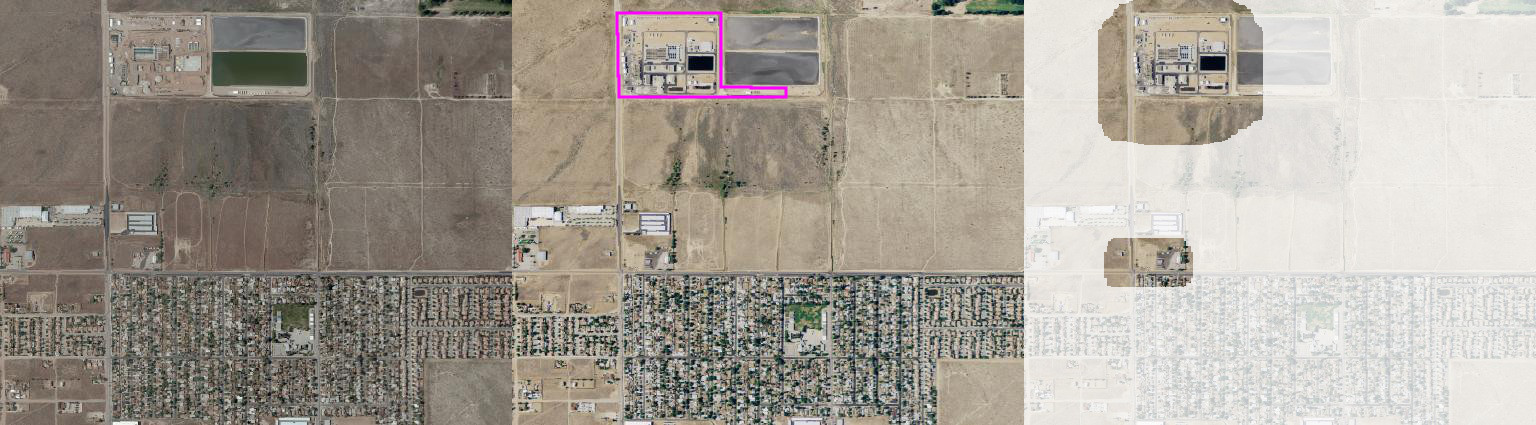}}
\hfill
\subfloat[]{\includegraphics[width=\textwidth]{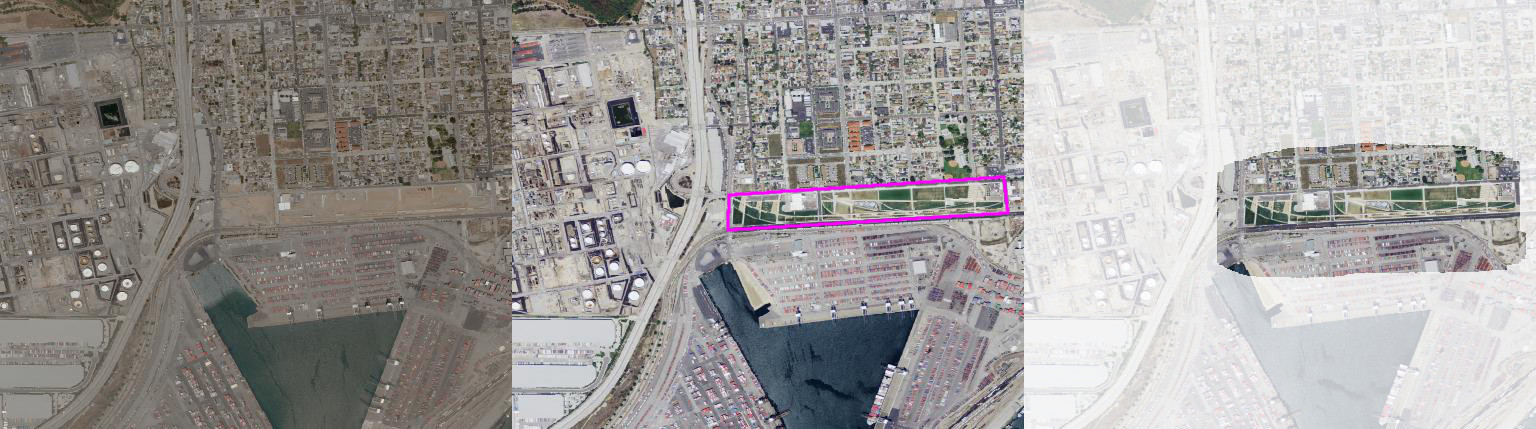}}
\hfill
\subfloat[]{\includegraphics[width=\textwidth]{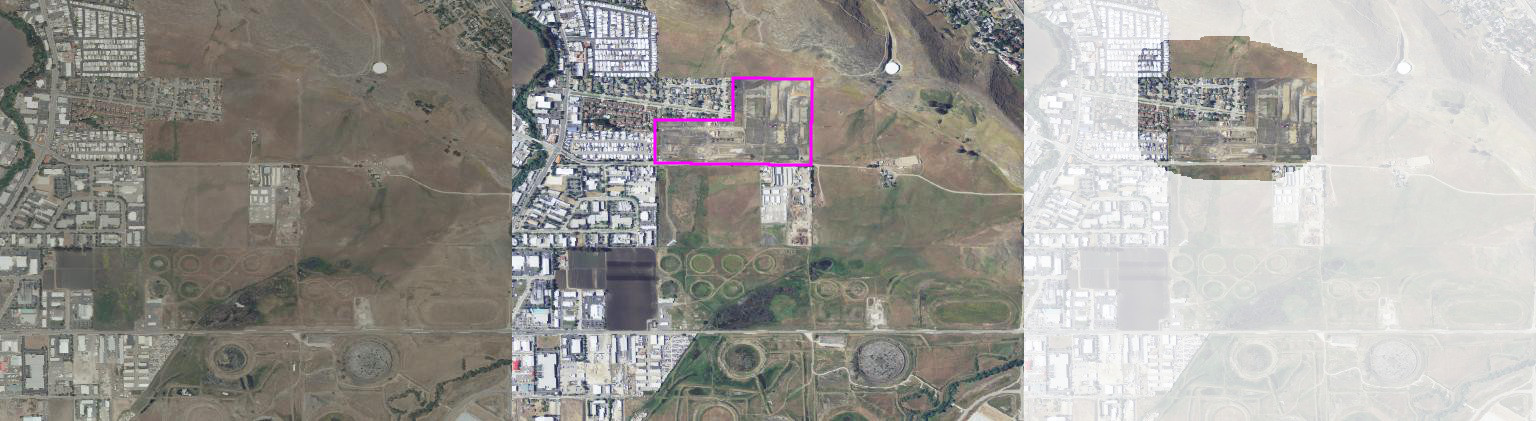}}
\caption{Examples of correct positive detections on the benchmark dataset, with threshold $\tilde\varepsilon = 10^{-4}$. Moving left to right, each triplet contains a scene in 2010, the scene in 2012 with labeled change region, and the change window proposed by the algorithm.  (a) Sports complex,  Los Angeles. (b) Wastewater treatment plant, Palmdale. (c) Waterfront park at Los Angeles Harbor. (d) Grading for a housing development, San Luis Obispo. The algorithm hones in on changes in large-scale, human-built structures despite complex background signals from differences in sensor calibration, in vegetation and desert ground cover, in spectral flux from the water, and from harbor activity.}
\label{fig:ExCorrect}
\end{figure}

\begin{figure}[!thpb]
\centering
\subfloat[]{\includegraphics[width=\textwidth]{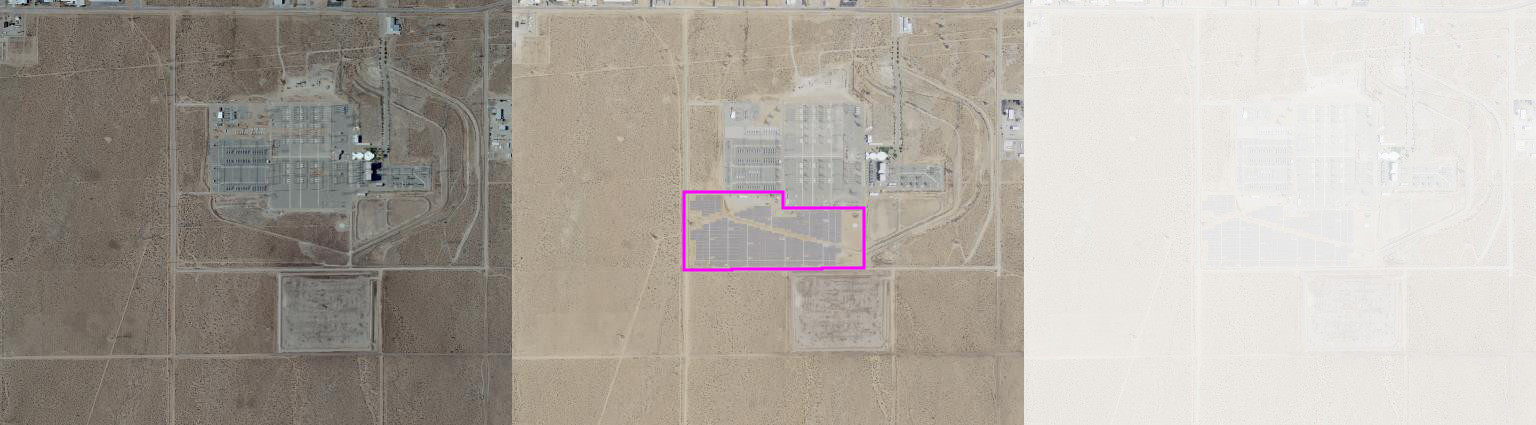}}
\hfill
\subfloat[]{\includegraphics[width=\textwidth]{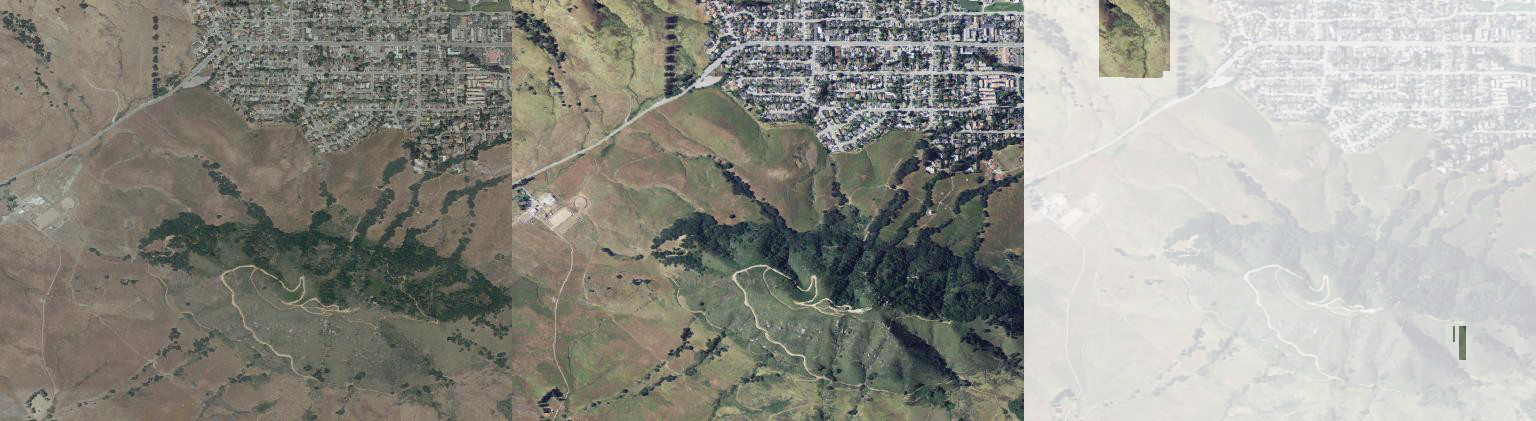}}
\hfill
\subfloat[]{\includegraphics[width=\textwidth]{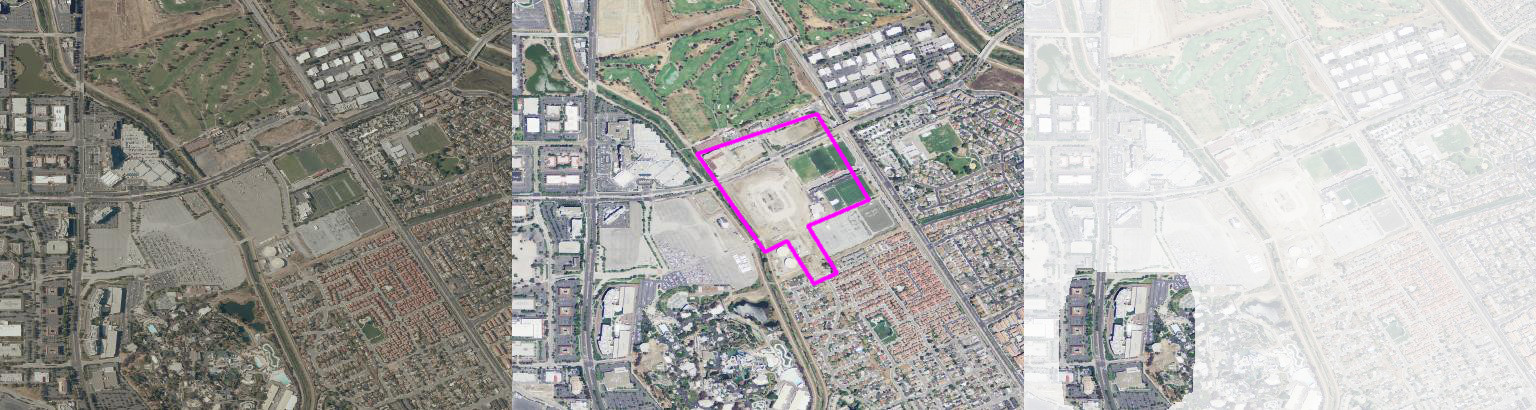}}
\hfill
\subfloat[]{\includegraphics[width=\textwidth]{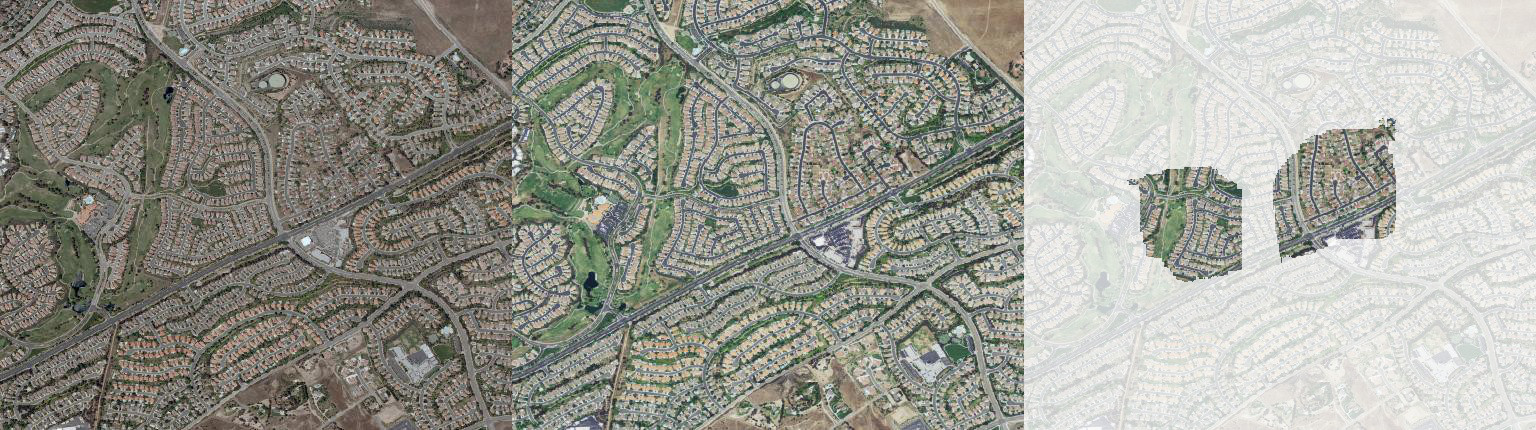}}
\caption{Fail cases from the benchmark dataset, with threshold $\tilde\varepsilon = 10^{-4}$.  Each triplet contains a scene from 2010, the scene in 2012 with (possible) labeled change region, and (possible) change window proposed by the algorithm. (a) Photovoltaic array outside Victorville: Missed because few total keypoints were extracted for this scene. (b) Outside San Luis Opisbo: A negative instance from the dataset. Differences in illumination of the folds in the landscape can be a source of false positive detections. (c) Initial construction of Levi's stadium, Santa Clara: A dual miss - the transition from parking lot to job site is relatively subtle, and the false detection centered on the tall buildings at lower-left may be due to differences in solar and viewing angles. (d) Housing developments, Temecula: Another false positive. Although we don't understand this case, speculatively we note the difference in radiance from the asphalt in the two images.}
\label{fig:ExIncorrect}
\end{figure}

Examples of successful detections and notable fail cases on the benchmark dataset are presented in Figs.~\ref{fig:ExCorrect} and~\ref{fig:ExIncorrect}. They were generated with the settings in Table~\ref{tab:params} and $\tilde\varepsilon = 10^{-4}$. As far as possible, we have tried to offer insights as to likely causes of misdetections, e.g., low match rates or clear differences in illumination. Despite the broad challenge of the task, the program makes credible, targeted assertions of change across a wide variety of scenes.

\section{Conclusions}

We have shown an ability to identify regions of large-scale construction against a background of diurnal and seasonal change.  Our strategy was to seek improved feature extraction and matching, which allows a neighborhood with many unmatched keypoints to register as a statistical anomaly against a a robust statistical model of matched keypoints on the image. KAZE features are the critical design element. They are sensitive to sharp edges and offer some invariance to non-rigid deformations and differences in illumination between images.  

Current prospects for application of the platform seem best in the filtering context mentioned above, where false positives can be tolerated as long as high-quality targets are also proposed.  We can tune the change probability threshold to the point that many quality positives are generated for every false detection, at least where positive and negative instances are balanced in the dataset. In a realistic filtering scenario, quality positives may, however, be relatively rare in the set of images. Along with direct applications, the filter functionality could aid in labeling data for future deep learning approaches to change detection.

In closing, we note that the spatial resolution of the imagery seems to impact the scale of objects most readily detected.  We experimented with native one-meter-resolution NAIP imagery and found that while average match rates decrease slightly, the algorithm performs in a manner qualitatively similar to that presented here. At one-meter resolution the algorithm captures construction of individual buildings.

\section*{Acknowledgments}

We are grateful for the support from The Earth Genome. We thank Robin Kraft and Dominika Blackappl for inspiration and technical support in the early phase of this work.

\ifCLASSOPTIONcaptionsoff
  \newpage
\fi
\bibliographystyle{IEEEtran}

\par\vfill\break
\advance\voffset by -2cm

\end{document}